# Feature Representation for Online Signature Verification

Mohsen Fayyaz, Mohammad Hajizadeh_Saffar, Mohammad Sabokrou and Mahmood Fathy

***Abstract*—Biometrics systems have been used in a wide range of applications and have improved people authentication. Signature verification is one of the most common biometric methods with techniques that employ various specifications of a signature. Recently, deep learning has achieved great success in many fields, such as image, sounds and text processing. In this paper, deep learning method has been used for feature extraction and feature selection, which has enormous impact on the accuracy of signature verification. This paper presents a method based on self-taught learning, in which a sparse autoencoder attempts to learn discriminative features of signatures from a large unlabeled signature dataset. Then, the features learned are employed to present users' signatures by creating a model for each user based on user genuine signatures. Finally, users' signatures are classified using a one-class classifier. The proposed method is independent on signature datasets thanks to self-taught learning. The features have been learned from 17,500 signatures (ATVS dataset) and verification process of the proposed system is evaluated on SVC2004 and SUSIG signature datasets, which contain genuine and skilled forgery signatures. The experimental results indicate significant error reduction and accuracy enhancement in comparison with state of the art counterparts.**

## I. Introduction

PEOPLE Authentication, has been known as an intrinsic part of social life. Recent years have seen a growing interest toward personal identity authentication. Increasing security requirements have placed biometrics at the center of a much attention. Biometric technology has become an important field in verifying people and has been used in people identification and authentication. The term biometrics refers to individual recognition based on a person's distinguishing characteristics [1]. In biometric systems, attributes do not have the disadvantages of token-based approaches that can be lost or stolen or knowledge-based approaches that can be forgotten. Therefore, biometric authentication systems have been used in a wide range of applications, such as; banking consumer verification, access control systems, etc.

M. Fayyaz and M. Hajizadeh are M.S. students at Malek-Ashtar University of Technology, Tehran, Iran (email: mohsen.fayyaz89@gmail.com, hajizadeh.m@gmail.com).

M. Sabokrou is Ph.D. student at Malek-Ashtar University of Technology, Tehran, Iran (email: sabokro@gmail.com).

M. Fathy is with School of Computer Engineering, Iran University of Science and Technology, Tehran, Iran (email: mahfathy@iust.ac.ir).

People recognition systems based on biometrics have two main categories [2]:

- Physiological biometrics are based on recognizing some physical part of the human body, such as fingerprint, retina, hand scan, etc.
- Behavioral biometrics are based on measuring some characteristics and behaviors of the human, such as handwritten signature, voice, etc.

Recognition refers to two different tasks: identification and verification. Identification specifies which user provides a given biometric parameter among a set of known users. Therefore, the input used for identification only contains genuine data. However, verification determines if the given biometric parameter is provided by a specific known user or is a forgery. Forgery consist of three types:

- Random forgery: Produced with no knowledge about the signature shape or signer's name
- Simple forgery: Produced by knowing only the name of the signer
- Skilled forgery: Produced by looking at the original signature sample

Person recognition has been applied by several biometric modalities, such as; fingerprint, iris, face, vein and signature [3]. Handwritten signature recognition is one of the most common techniques to recognize the identity of a person. However, when dealing with signatures, most of the proposed systems focus on verification rather than identification because of daily usage of signature verification systems [4].

There are two types of signature verification: Offline (static) verification and Online (dynamic) verification. In the offline setting, we have the shape of the signature by capturing or scanning them from papers and the system must extract features from the picture of the signature. Therefore, in offline verification system, input data contains x-y coordinates of signatures. However, in the online setting, the system uses devices for capturing additional information while the user is signing [5]. Online signatures have extra information for extraction such as time, pressure, pen up and down, azimuth, etc.

Two types of features can be extracted from a signature [1] (Figure 1):

- Function-Features: The signature is characterized in terms of a time function whose values constitute the feature set, such as position, velocity, pressure, etc.

- ❖ Parameter-Features: The signature is characterized as a vector of elements each representing a value of a feature. Parameters are generally classified into two main categories: local and global. Local features are so-called because of their relation to each point of the signature, such as height or width ratio of the stroke, stroke orientation, pixel density, etc. Global features are so-called because of their relation to the whole of the signature and signing process, such as total time, average pressure, average speed, etc.

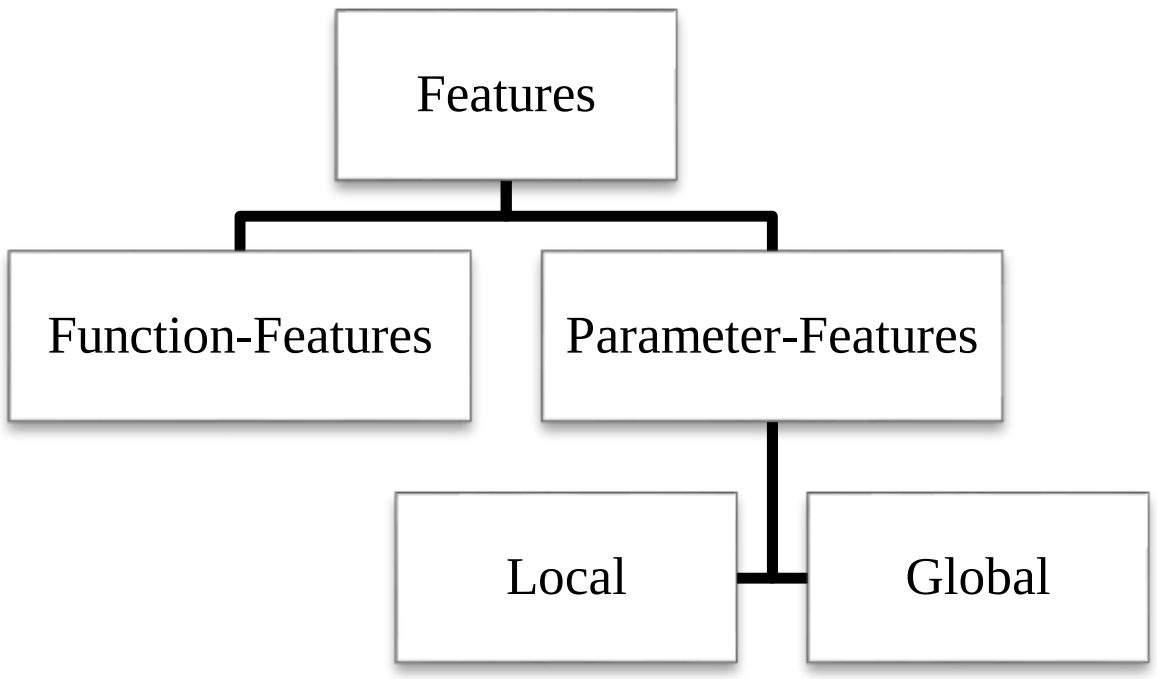

Figure 1 Feature categories

The Verification approaches can be described in three categories [1]:

- ❖ Template Matching: A questioned sample is matched against templates of signatures, such as Dynamic Time Wrapping (DTW) [4-6], Euclidian distance.
- ❖ Statistical: In this approach, distance-based classifiers can be considered, such as Neural Networks [7], Hidden Markov Models (HMM) [4, 8]..
- ❖ Structural: This approach is related to structural representations of signatures and compared through graph or tree matching techniques [9].

In this paper, a signature verification system has been proposed based on deep learning. The sparse linear autoencoder has been implemented to learn the signature model of each user by learning features based on an unsupervised self-taught method. Furthermore, one-class classifier has been used for classifying test signatures.

This paper is organized as follows: Section II presents a brief description of related work in the field of online signature verification. Section III introduces the adopted methodology while section IV presents the proposed system. Experimental results and their comparisons have been described in section V. Finally, section VI presents the conclusion for this paper and suggestions for future work.

## II. Related Work

There is an extensive literature in the field of online signature verification. Most recent approaches have been described in [1, 2, 10]. The process of signature verification is usually divided into three phases:

### A. *Preprocessing*

The signature dataset must take some preprocesses since there is no guarantee that different signatures of one user will always be the same. Several processes have been proposed for this phase, which generally consist of smoothing, rotation and normalization.

Cubic splines can be employed for smoothing purposes to solve the jaggedness in the signatures. Signatures can become rotation-invariant by rotating each signature based on orthogonal regression (Eq.1) [5].

$$\bar{\theta} = tg^{-1}\left(\frac{{s_y}^2 - {s_x}^2 + \sqrt{({s_y}^2 - {s_x}^2)^2 + 4*({cov_{(x,y)}}^2)}}{2*cov(x,y)}\right) \quad (1)$$

Where $s_x$ and $s_y$ are variance and $cov(x,y)$ is covariance of the horizontal and vertical components.

The signatures of one person must have the same size for better performance. The horizontal and vertical components of the signatures can be normalized to make a standard size of signature (Eq. 2, 3) [6].

$$x_n = \frac{x - \min(x)}{\max(x) - \min(x)} * 100 \quad (2)$$

$$y_n = \frac{y - \min(y)}{\max(y) - \min(y)} * 100 \quad (3)$$

Where $x$ and $y$ are original and $x_n$ and $y_n$ denote the normalized coordinates.

### B. *Feature Extraction*

Feature selection and feature extraction play an important role in verification systems. Many studies have done in the field of feature selection to choose the best set of features for extraction. List of common features have been described in Table 1 [5].

Table 1 List of common features

| **List of common features** | |
|---|---|
| **#** | **Description** |
| 1 | Coordinate $x(t)$ |
| 2 | Coordinate $y(t)$ |
| 3 | Pressure $p(t)$ |
| 4 | Time stamp |
| 5 | Absolute position, $r(t) = \sqrt{x^2(t) + y^2(t)}$ |
| 6 | Velocity in x, $v_x(t)$ |
| 7 | Velocity in y, $v_y(t)$ |
| 8 | Absolute velocity, $v(t) = \sqrt{{v_x}^2(t) + {v_y}^2(t)}$ |
| 9 | Velocity of r(t), $v_r(t)$ |
| 10 | Acceleration in x, $a_x(t)$ |
| 11 | Acceleration in y, $a_y(t)$ |
| 12 | Absolute acceleration, $a(t) = \sqrt{{a_x}^2(t) + {a_y}^2(t)}$ |

Furthermore, some non-common features have been described in other papers [6, 9, 11-15]. Recently, some biometric authentication systems for face, iris and fingerprint have been proposed based on deep neural networks which used autoencoders for feature extraction phase [3, 16].

### *C. Classification*

After the feature extraction phase, the system must learn the features extracted from reference signature. For classification phase, each signature must be compared against reference signatures and the difference between features of test signature and reference signatures would be calculated. By having the distances between test and reference signatures, the system can decide to accept or reject the test signature.

There are different options for distance calculation such as $d_{min/max}$ which is minimum/maximum distance between a signature and the patterns of the reference set, and $d_{central}$ which is the distance between a signature and the center of mass of the reference set [17]. One of the important parameter in verification system is the threshold value for accepting or rejecting a signature. Consequently, choosing the best threshold is a crucial step. There are two types of thresholds: global and local. In global threshold, the system will choose one threshold value for all users. On the other hand, for local threshold, the system must choose one threshold per user so that, this approach could lead to a better result [17].

As mentioned, the signature recognition problem is an abstract concept, which comprises signature identification and signature verification. In daily usage of authenticating systems such as banking systems, handwritten signature of users have been used to verify the identity of official documents. In these sorts of problems, the main goal is verifying whether a signature belongs to one identified person or not. In contrast with multi-class classifiers, the aim for one-class classifiers is distinguishing one type of class (target) from other classes (outlier). Thus, For classifying a signature as genuine or forgery, one-class classifiers have been commonly used [17] to divide the set into two categories: target and outlier (Figure 2).

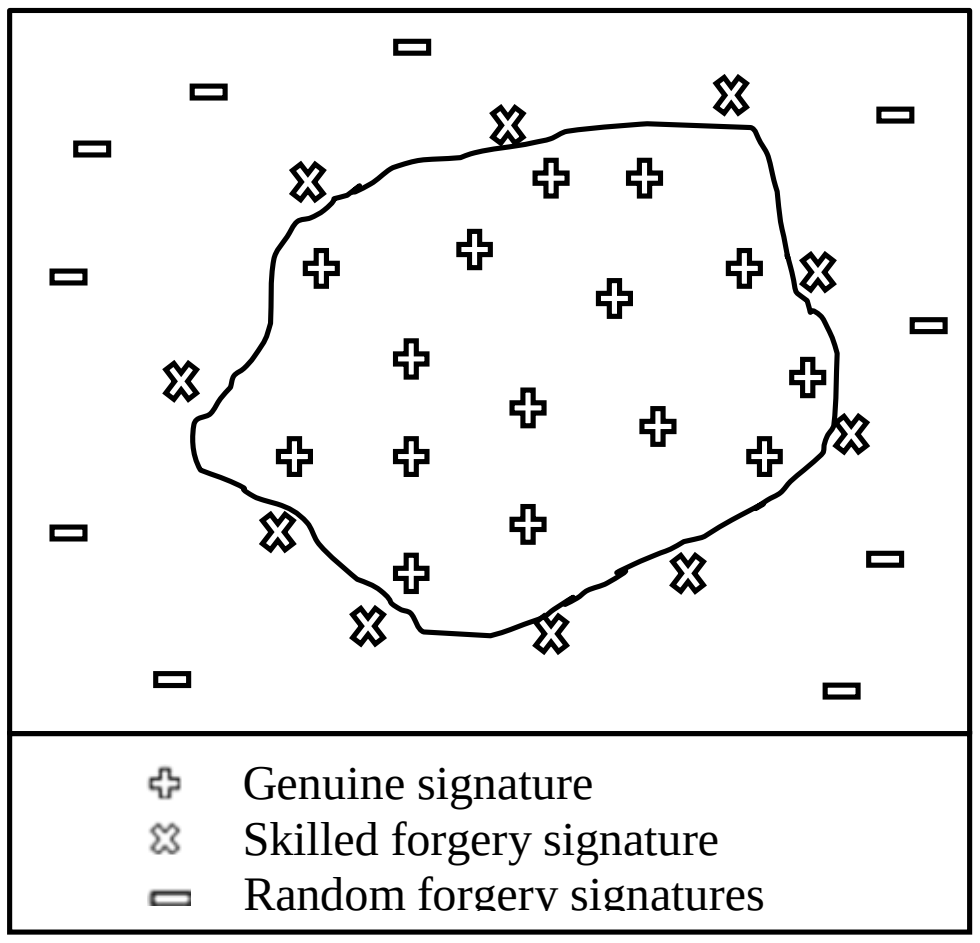

Figure 2 Example of signature model for each user

Jain and Gangrade [7] proposed a system by using angle, energy and chain code features to diffrentiate the signatures. In this approach, a Neural Network has been applied for classification.

Faundez-Zanuy [4] studied four pattern recognition algorithms for online signature recognition: Vector Quantization (VQ), Nearest Neighbor, DTW and HMM. The author proposed two methods based on VQ and Nearest Neighbor.

Rashidi, et al. [5] evaluated 19 dynamic features viewpoint classification error and discrimination capability between genuine and forgery signatures. They used a modified distance of DTW for improving performance of verification phase.

Ansari, et al. [6] presented an online signature verification system based on fuzzy modelling. The point of geometric extrema has been chossen for signature segmentation and a minimum distance alignment between samples has been made by DTW techniques. Dynamic features have been converted to a fuzzy model and a user-dependent threshold used for classification.

Barkoula, et al. [9] studied the signatures Turning Angle Sequence (TAS), the Turning Angle Scale Space (TASS) representations, and their application to online signature verification. In the matching stage, the authors have employed a variation of the longest common sub-sequence matching technique.

Yahyatabar, et al. [11] proposed a method based on efficient features defined in persian signatures. A combination of shape based and dynamic extracted features has been applied and a SVM has been used for classification phase.

Alhaddad, et al. [12] explored a new technique by combining back-propagation Neural Network (BPNN) and the probabilistic model. BPNN has been used for local features classification, while probabilistic model has been used to classify global features.

Mohammadi and Faez [13] proposed a method based on the correspondence between important points in the direction of wrap for the time signal provided to maximize the distinction between the genuine and forged signatures.

Napa and Memon [14] Presented a simple and effective method for signature verification in which an online signature is represented with a discriminative feature vector derived from attributes of several histograms that can be computed in linear time. For testing phase, the authors proposed a method on finger drawn signatures on touch devices by collecting a dataset from an uncontrolled environment and over multiple sessions.

Souza, et al. [17] proposed an off-line signature verification system, which uses a combination of five distance measurements, such as, furthest, nearest, template and central using four operations: product, mean, maximum, and minimum as a feature vector.

Fallah, et al. [18] presented a new signature verification system based on Mellin transform. The features have been extracted by Mel Frequency Cepstral Coefficient (MFCC). Neural Network with multi-layer perception architecture and linear classifier in conjuction with Principal Component Analysis (PCA) have used for classification.

Iranmanesh, et al. [19] proposed a verification system by using multi-layer perceptron (MLP) on a subset of PCA features. This approach used a feature selection method on the information that has been discarded by PCA, which significantly reduced the error rate.

Cpałka, et al. [20] explored a new method by using area partitioning of high and low speed of the signature and high and low pen's pressure. The template for each partition has been generated and by calculating the distance between signatures and template in each partition, a fuzzy classification has been implemented to classify the signatures.

Lopez-Garcia, et al. [21] presented a signature verification system implemented on an embedded system. In this approach, a template for each user has been generated and a DTW algorithm has been used for distance calculation. Finally, the features extracted and passed through a Gaussian Mixure Model (GMM) to calculate the similarity between the test signature and the generated template.

Gruber, et al. [22] proposed a technique based on Longest Common Subsequences (LCSS) detection. Authors have used a LCSS kernel of SVM for classifying the similarity of signature time series.

## III. METHODOLOGY

Deep learning (Feature Learning or Representation Learning) is a new era of machine learning which aims to learn the high-level features from raw data to achieve a better performance in classification tasks. Deep learning is part of a field of machine learning methods based on learning representation of data [23].

Raw data (e.g. an image) can be represented in many ways by using diverse handcrafted features. Feature learning tries to learn discriminative features autonomously which is one of its advantages. The other advantage of feature learning is that the feature learning process can be completely unsupervised. One of the goals of deep learning is hierarchical feature extraction. For achieving that goal, feature learning tries to learn a new representation of the input data which is the observed data and continue learning new representations of previously learned features at each level, which are able to reconstruct the original data.

One of the scopes of machine learning, which plays a key role in deep learning, is self-taught learning. The main promise of self-taught learning is using unlabeled data in supervised classification tasks [24]. The key point of such algorithms is that unlabeled data are not supposed to follow the same class labels. Indeed, unlabeled data are exploited to teach the system recognizing patterns or relations for the supervised learning task. In summary, self-taught learning learns a concise, higher-level feature representation of the raw data using unlabeled data. Having a concise high level feature representation brings us an easier classification task by having features that are more significant [24].

### A. Autoencoder

One of the unsupervised learning methods is the autoencoder algorithm. Autoencoder is an unsupervised learning architecture used to pre-train deep networks. There is one kind of autoencoder algorithm, which is based on multi-layer perceptron neural networks. In contrary to traditional neural networks, MLP based autoencoders are unsupervised learning algorithms which try learning weights of each layer to set the output values to be equal to the inputs for the neural network (Figure 3).

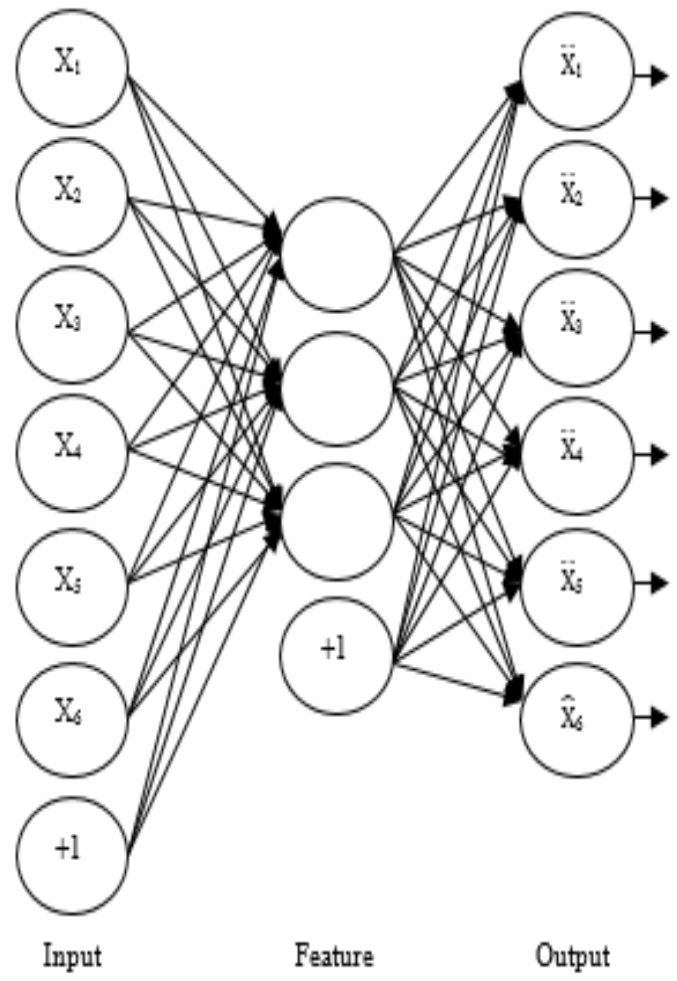

Figure 3 Architecture of Autoencoder

Suppose $x \in \mathbb{R}$ is the set of input features. To learn features from input features, the basic autoencoder with regularization term to prevent over-fitting, attempts reconstructing input features by minimizing following cost function (Eq. 4):

$$J(W,b) = \arg\min_{W,b} \frac{1}{m} \sum_i \left\| h_{w,b}\left(x^{(i)}\right) - x^{(i)} \right\|^2 + \lambda \sum_l \sum_{i,j} (W_{i,j}^l)^2 \quad (4)$$

Where $W \in \mathbb{R}$ is weight matrix mapping nodes of each layer to next layer nodes, and $b \in \mathbb{R}$ is a bias vector.

The cost function of autoencoder mentioned in (Eq. 4) only focuses on the differences between input and output data of autoencoder. This brings us a network with the ability of representing raw data with learned feature without any guarantee of having sparse represented features, which plays a key role in classification task. In order to learn features that are more effective and having a sparser dataset of represented features, the sparsity constraint can impose on the autoencoder network. The objective function is as follows (Eq. 5-7):

$$J_{Sparse}(W,b) = J(W,b) + \beta \sum_i KL(\rho || \hat{\rho}_j) \quad (5)$$

$$KL(\rho || \hat{\rho}_j) = \rho log \frac{\rho}{\hat{\rho}_j} + (1-\rho) log \frac{1-\rho}{1-\hat{\rho}_j} \quad (6)$$

$$\hat{\rho}_j = \frac{1}{m} \sum_i \left[ a_j^2 (x^{(i)}) \right] \quad (7)$$

Where $KL(\rho || \hat{\rho}_j)$ is the Kullback-Leibler ($KL$) divergence between a Bernoulli random variable with mean $\rho$ and a Bernoulli random variable with mean $\hat{\rho}_j$, which is the average activation of hidden unit $j$. The notation summary of equation 4-7 is described in Table 2.

Table 2 Autoencoder cost function notation summary

| Autoencoder cost function (Eq. 4-7) notation summary | |
|---|---|
| **Symbol** | **Description** |
| $x$ | Input features for a training example |
| $y$ | Output/Target values. $y$ is a vector. In the case of an autoencoder, $y = x$ |
| $x^{(i)}$ | The $i$-th training example |
| $W$ | The parameter associated with the connection between units of layers |
| $b$ | The bias term associated with the connection between two layers |
| $\rho$ | Sparsity parameter, which specifies the desired level of sparsity |
| $\hat{\rho}_i$ | The average activation of hidden unit $i$ (in the sparse autoencoder) |
| $\beta$ | Weight of the sparsity penalty term (in the sparse autoencoder objective) |
| $\lambda$ | Weight decay parameter |

A sparse autoencoder model can effectively realize feature extraction and dimension reduction of the input data, which play a vital role in classification tasks [16].

### *B. Convolution and Pooling*

Raw input data are usually stationary. It means that the statistics of randomly selected parts of the data are the same. This characteristic shows that not all the features are useful. It is obvious that having more features results in increasing the computational complexity especially in a classification task. In order to avoid high computational complexity, redundant data have been neglected by picking up random patches of raw data and convolving them. After obtaining convolved features, pooling method can be exploited in order to obtain pooled convolved features. These pooled features can be used for classification task (Figure 4).

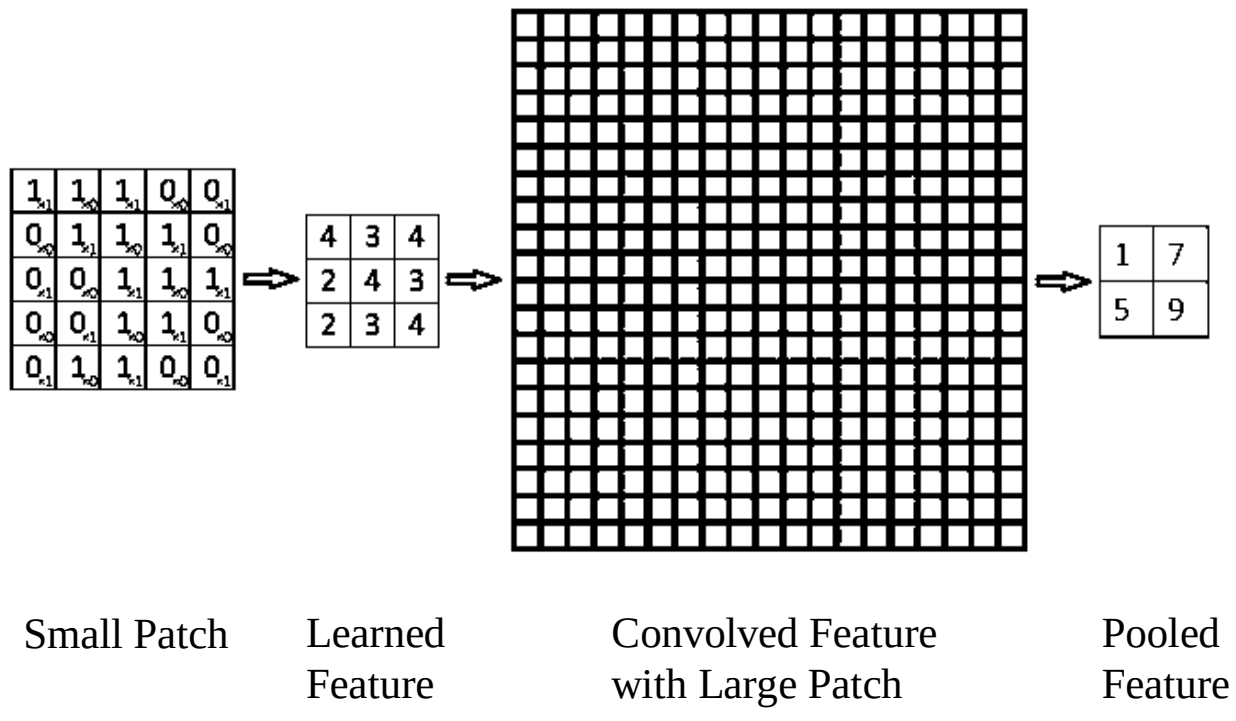

Figure 4 Convolution and pooling example

## IV. Proposed System

One of the important problems in signature verification is choosing features due to diverse difficulties in signature verification, such as differences between same user signatures, different circumstances of signing, various shapes of signatures, etc. Among these, exploiting an unsupervised feature learning method results in system compatibility improvement with various types of signatures and automatic feature selecting from signatures. The proposed signature verification system comprises three steps: Feature learning, One-class classification and Verification (Figure 5).

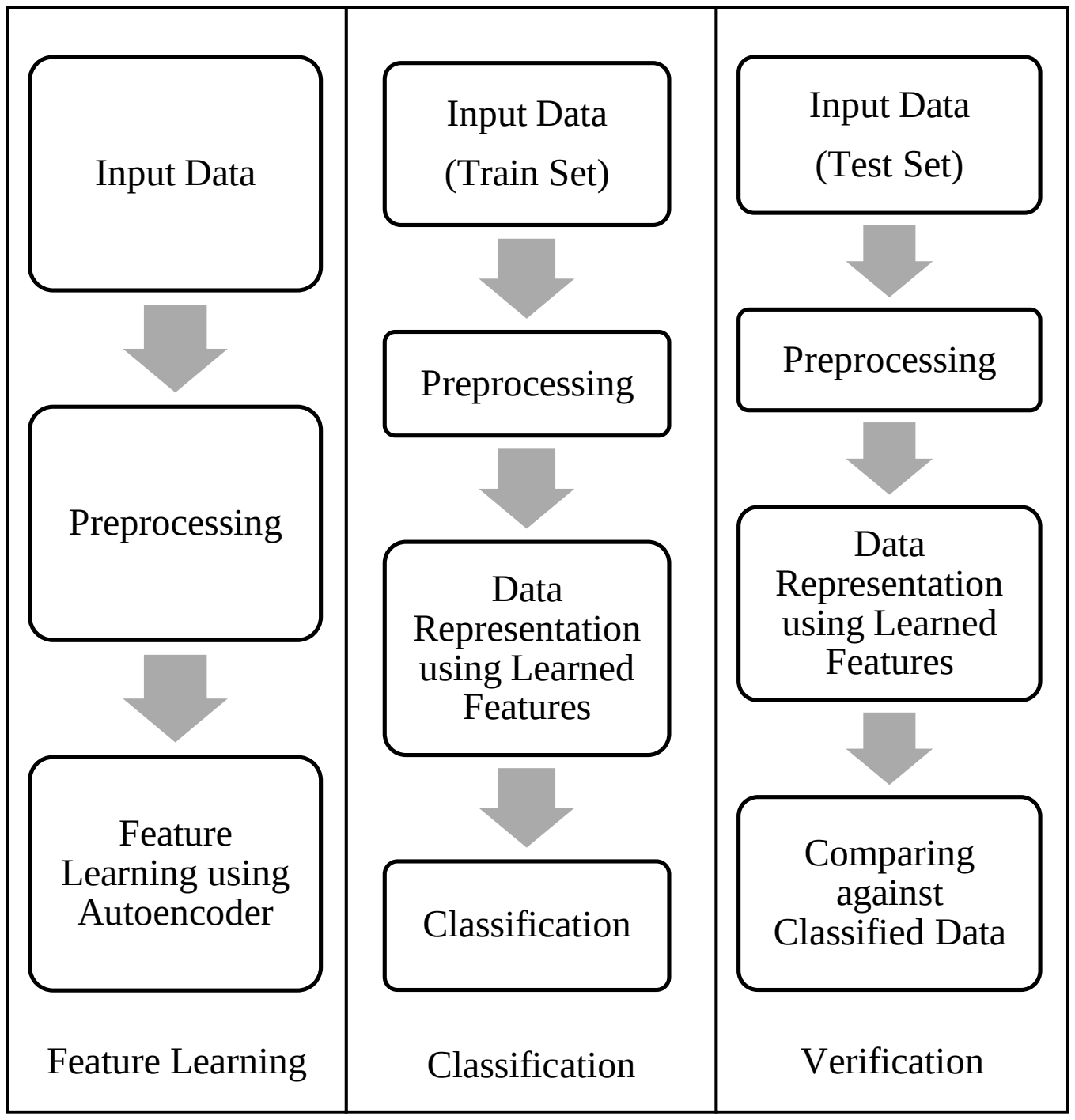

Figure 5 Proposed system architecture

In the first step, named feature learning, features are learned by the autoencoder. In this step, an unlabeled dataset, which is discretized from train and test datasets, is used based on self-taught method. In classification step, a reference model of the system is built using classified represented data from users' reference signatures. These two steps are parts of the system training section [14]. Finally, in verification step, which is system-working section, new unknown signatures are compared against the system reference model (classified data) to be verified. There are three principal phases among described steps, which are preprocessing, feature learning using autoencoder, and classification. These phases are explained as follows:

### *A. Preprocessing*

As mentioned, in the preprocessing phase, the first step is normalizing size of the signature. This aim can be achieved by scaling the signature size. At the next step, the mean of the data must become equal to zero for data normalization.

Signatures data in databases are based on time, pressure, pen up/down, etc. in x/y positions. To make representation become similar to reality, points of signatures have been continued. This object achieved by using time of the points to observe the sequence of data and pen up /down to check if the pen has gone up, the point must be separated from the next one. Finally, signatures have been represented base on two layer: pressure and time (Figure 6).

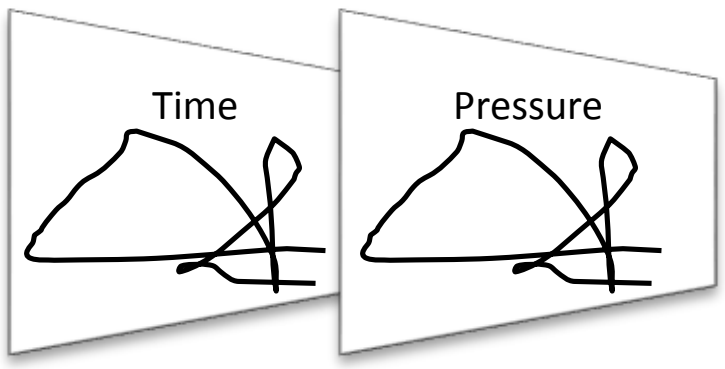

Figure 6 Illustration of input signature

Principal Component Analysis (PCA) is an algorithm that reduces dimensions of signature data and can be used to significantly speed up unsupervised feature learning algorithm. Since the system is trained based on signature images, adjacent pixel values are highly correlated. Whitening can make the input less redundant, the features become less correlated with each other and the features all become the same variance. Therefore, these two algorithms have used to reduce the dimension.

### *B. Feature Learning using Autoencoder*

For learning features from signatures, a linear autoencoder with sparsity have been used. The signature has been set for input and output and autoencoder has been checked to maps input to output. This autoencoder has been designed based on gradient descent.

Unsupervised learning algorithms have high computational cost. In order to increase performance of learning phase, raw data (large patch of a signature) has been divided into small patches and have been used in feature learning phase as input. Then learned features have been convolved with large patch. After obtaining features using convolution, mean pooling method has been exploited in order to obtain pooled convolved features. These pooled features have been used for classification.

### *C. Classification*

The significant issues of classification in this type of problems are differences between same user's signatures, diverse circumstances of signing, low amount of signature samples, and forgery signatures. For resolving such issues, selecting an appropriate classifier is very important.

The one-class classifier in the proposed system has a target class, which is class of the user whose signature is being compared with input signature, and the outlier class is other users' sample signatures. As a result, the classifier must create a model of target class for each user.

## V. Experimental Results

In the evaluation process of proposed approach, test signatures have been comprised by comparing their features against reference signatures. In this section, short description of benchmarks and evaluation parameters have been described. In addition, three steps of the proposed system are explained.

### *A. Benchmarks*

For evaluation of the proposed approach, three public datasets have been used which are SVC2004 [25], SUSIG [26] and ATVS [27, 28]. The structure of the mentioned datasets have been explained as follows:

#### *1) SVC2004*[1]

SVC2004 is the first international signature verification competition. The aim of holding SVC2004 competition was allowing researchers to evaluate the performance of their signature verification methods based on benchmark datasets and benchmarking rules that resulted in creating a benchmark dataset named SVC2004.

SVC2004 main database has 100 sets of signature data. SVC2004 public database, which has been released before the competition, consists of 40 signature sets. Each set includes 20 genuine signatures of one signature contributor and 20 skilled forgeries of at least four other contributors (Figure 7).

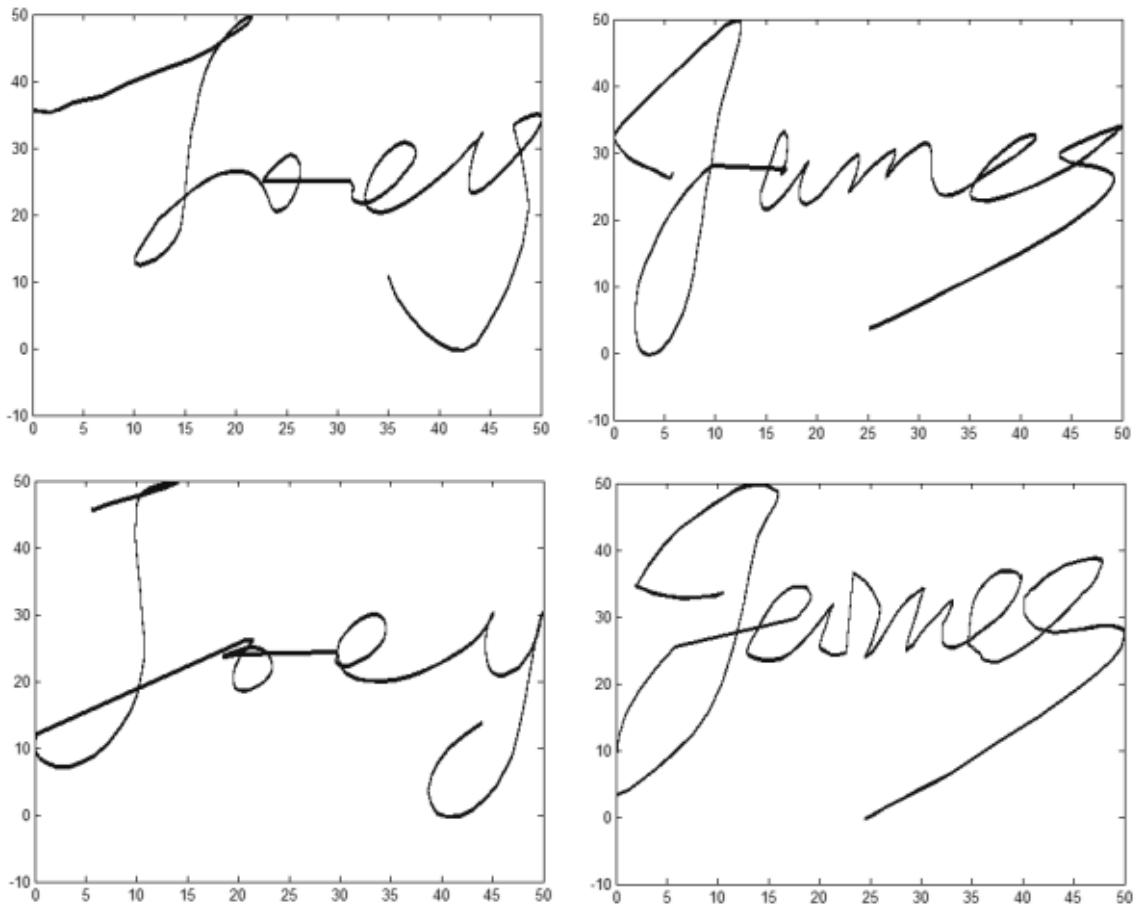

Figure 7 Examples of Genuine (first row) and Forgery (second row) signatures in SVC2004 database

In data collection process of the signature sets, contributors were asked not to use their real signatures for privacy reasons. On the other hand, made-up signatures are shortcoming of this database, which will result in having higher variance and higher error rates. For decreasing effect of the mentioned problem, contributors were reminded that, not only should their signatures have spatial consistency in signature shape but should have temporal consistency of dynamic features as well. Contributors were asked to contribute 20 genuine signatures in two sessions in two weeks. At least four other contributors forged the skilled forgeries for each contributor's signature.

In SVS2004 database, each signature includes a sequence of points, which contains X, Y coordinates, time and pen up/down, azimuth, altitude and pressure.

#### *2) SUSIG*[2]

Sabanci University Signature database (SUSIG) is a database of online signatures, which aim is overcoming some of the shortcomings of its contemporary databases.

The SUSIG database consist of two subcorpora, which are visual and blind. In both subcorpora, contributors used their real signatures for creating genuine signatures sets, which is one of this database advantages in contrary to SVC2004 database (Figure 8).

[1] Available at http://www.cse.ust.hk/svc2004/download.html
[2] Available at http://biometrics.sabanciuniv.edu/susig.html

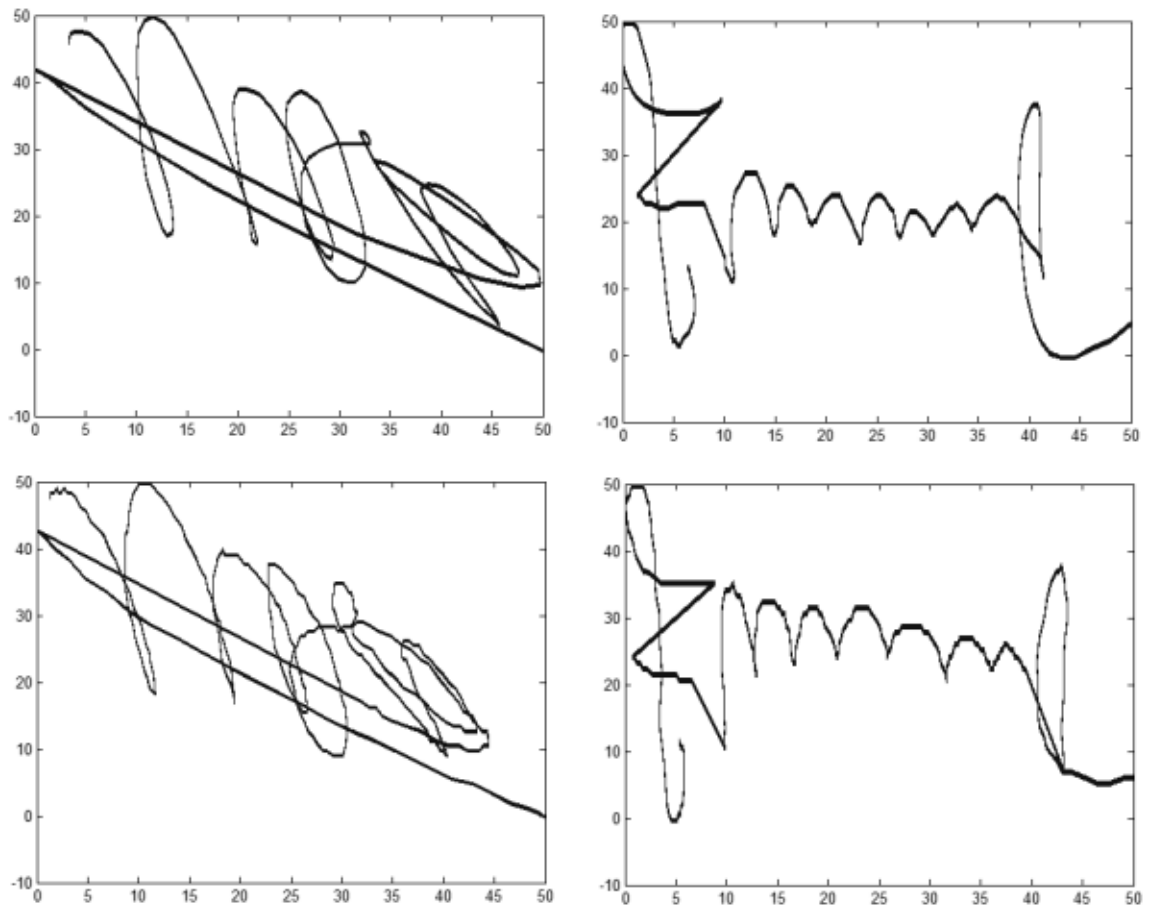

Figure 8 Examples of Genuine (first row) and Forgery (second row) signatures in SUSIG database

In blind subcorpus data collection process, collection has been done on a tablet without visual feedback. It consists of signatures of 100 contributors. First group of 30 contributors provided eight genuine signatures, while the other 70 contributors provided 10 genuine signatures each. For providing forgery signatures, forgers were shown the genuine signatures' drawing replay several times. After training well, forgers supplied ten forgeries for each set of contributors' genuine signatures. Additionally, there is a separate ten-person validation set with ten genuine and ten forged signatures per person.

In visual subcorpus data collection process, collection has been done on a tablet with a LCD, which provided visual feedback to the contributors while they were signing signatures. Visual subcorpus data were collected in two separate sessions. Each contributor has provided 20 samples of his/her signature. In this database, in the visual subcorpus, there are two types of forgery signatures: skilled forgeries and highly skilled forgeries. For providing skilled forgeries, contributors were shown the genuine signatures' drawing like the blind subcorpus. Each forger was asked to provide five forgeries of the signature. For providing highly skilled forgeries, the replay of the reference signature shown on both a monitor in front of forgers and the LCD screen of the tablet which provided forgers the ability of tracing the reference signature signing process. Like the normal skilled forgeries, forgers have forged five highly skilled forgeries for each set of genuine signatures. In summary, 20 genuine signatures and five skilled and five highly skilled forgeries were collected for each person in the subcorpus. Additionally, there is a separate 10 person-validation set with 10 genuine and 10 forged signatures per person acquired in a single session for tuning system parameters.

*3) ATVS*[3]

All two mentioned databases (SVC2004 and SUSIG) are human made database. Although they have advantages, such as, having real signature of a human, and having real world situations for sampling, they have restrictions, which are limited amount of data, privacy issues, subdued to legal aspects. Synthetic signature databases are solution of this problem. They are not restricted to limitations mentioned above. However, they miss the advantage of having real world situations, and real human signature. In spite of suffering from such problems, synthetic databases have had good approaches to simulation of real signatures, which involves the effect of real situation of sampling. ATVS database is one of the synthetic databases (Figure 9).

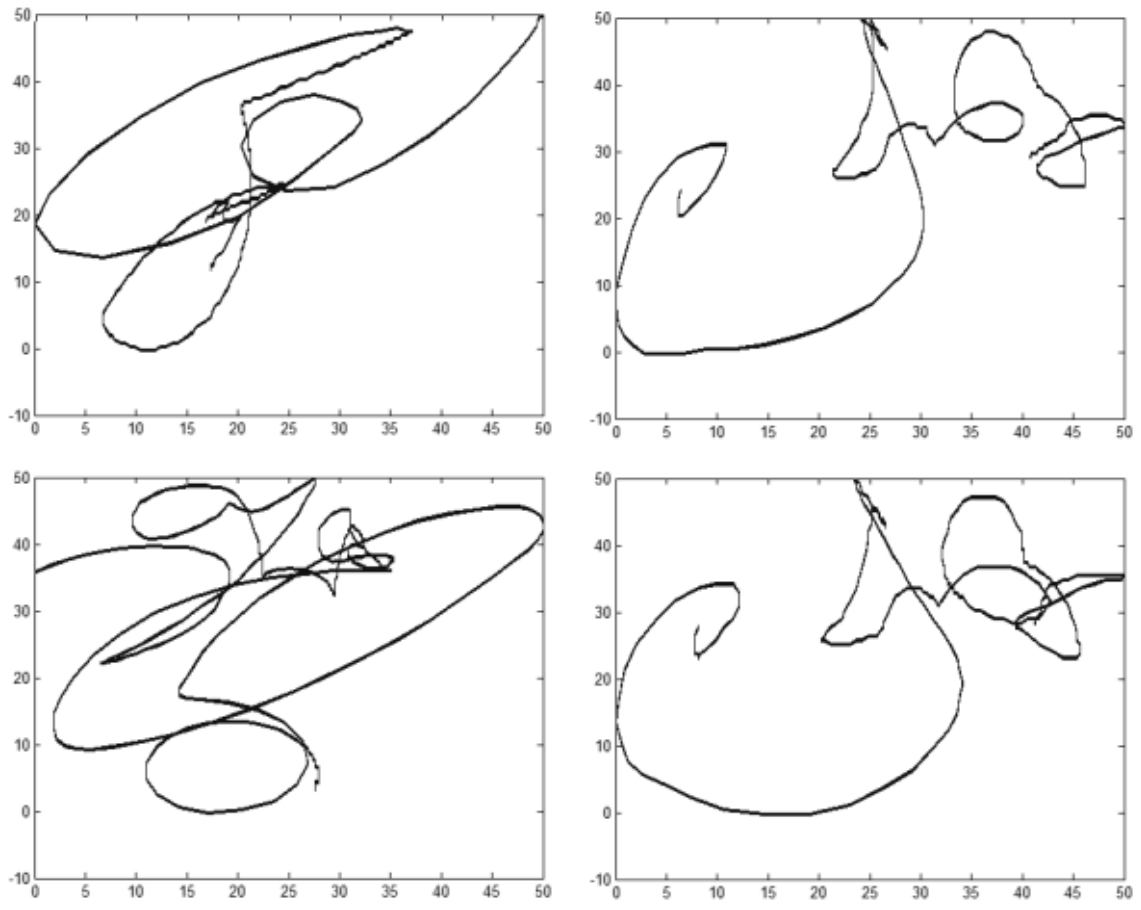

Figure 9 Examples of Genuine (first row) and Forgery (second row) signatures in ATVS database

The artificial samples produced by ATVS follow the pattern of the western signatures, which are left-to-right concatenated handwritten signature [27]. ATVS signature generation contains two steps: First, a master signature corresponding to a synthetic individual is produced using a generative model based on information obtained. This information has been acquired by analyzing real signatures using a spectral analysis approach and the kinematic theory of rapid human movements. Second, various samples of the same synthetic subject are created using the master signature.

ATVS has two parts, named "direct modification of the time functions" and "modification of the sigma-lognormal parameters (LN-Parameters)".

In direct modification of the time functions, time sequence of the reference signature have been modified according to a model simulating the distortions introduced by a given channel.

In Modification of the Sigma-Lognormal Parameters, authors have decomposed the velocity function v, derived from the coordinate functions x and y, into simple strokes. Each stroke is used with different velocity functions to set the Sigma-Lognormal parameters.

In summary, ATVS especially the ATVS-SSig have two types of data. In Modification time functions, as described, the time functions of the master signature is changed to generate the duplicated samples [27]. In modification LN-Parameters, duplicated samples are generated modifying the lognormal parameters of the master. Both methods use 25 signatures from 350 users. Of the 25 signatures, the first five follow an intra-session variability and the next 20 follow an inter-session variability.

[3] Available at http://atvs.ii.uam.es/databases.jsp

### B. Evaluation Parameters

Different parameters have been used in verification systems. In the following, a short description of most commonly used parameters have been summarized.

1) Receiver Operating Characteristic (ROC) Curve

A one-class classifier can be evaluated based on small fraction false negative (false reject rate) and false positive (false accept rate). ROC curve shows how the fraction false positive varies for varying fraction false negative. Traditionally the fraction true positive is plotted versus the fraction false positive. The smaller these fractions are, the more this one-class classifier is to be preferred.

*2) Equal Error rate (EER)*

If a line connects the points (1, 0) and (0, 1) in the ROC curve of a classifier, EER can be defined such that false positive and false negative fractions are equal. This parameter is a simple way to compare system accuracies. The smaller the EER rate is, the more accurate the system is.

*3) Area Under the ROC Curve (AUC)*

AUC is one way to summarize an ROC curve in a single number. This integrates the fraction true positive over varying thresholds (or equivalently, varying fraction false positive). Higher values indicate a better separation between target and outlier objects.

### C. Feature Learning

In feature learning phase, a methodology has been set to learn features based on a signatures set except of test and train sets. Therefore, all of the signatures in ATVS database have been used for feature extraction using autoencoder (Figure 10). The autoencoder comprises one hidden layer with 2000 nodes and the limited Broyden–Fletcher–Goldfarb–Shanno algorithm (L-BFGS) method with 700 iteration for minimization function.

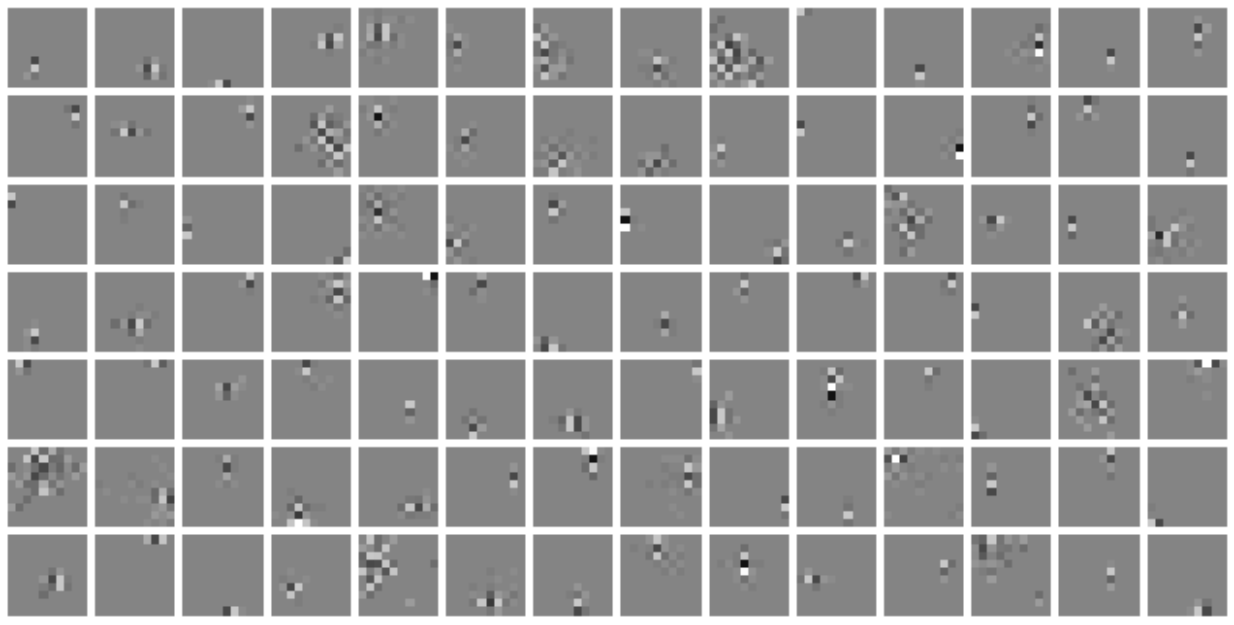

Figure 10 Illustration of features that were learned using autencoder

### D. Classification and Verification

In this phase, SVC2004 and SUSIG databases have been used for a K-Fold Cross-Validation process that has been implemented to categorize train and test signature groups. Several experiments have been done to achieve the best values for system parameters.

The size of hidden layer and iteration value have been selected based on an experiment on auto-encoder with hidden size of 500, 1000, 1500, 2000, 2500 and 3000 nodes in which the iteration value was set from 100 to 700. EER and AUC results for SVC and SUSIG databases have been shown in Table 3 and Table 4, respectively.

The results shown a decrement in EER and an increment in AUC rate while facing iteration value increment. Due to change mitigation in more than 700 iterations, the iteration value has been set to 700. Although for hidden size parameter, the rate of enhancement of EER and AUC rates decreased for hidden sizes larger than 2000 while computational costs increased and had been prone to over fitting and curse of dimensionality. Finally, the size 2000 has been selected because of its computational efficiency and appropriate accuracy.

Table 3 EER Experiment results with different hidden size for SVC2004 and SUSIG

| Iteration | 100 | | 200 | | 300 | | 400 | | 500 | | 600 | | 700 | |
|---|---|---|---|---|---|---|---|---|---|---|---|---|---|---|
| Hidden size | SVC | SUSIG | SVC | SUSIG | SVC | SUSIG | SVC | SUSIG | SVC | SUSIG | SVC | SUSIG | SVC | SUSIG |
| 500 | 1.7 | 5.02 | 1.65 | 4.94 | 1.60 | 4.77 | 1.60 | 4.74 | 1.55 | 4.57 | 1.45 | 4.07 | 1.03 | 3.23 |
| 1000 | 1.25 | 4.90 | 1.14 | 4.57 | 1.14 | 4.24 | 1.08 | 3.72 | 1.06 | 3.72 | 1.03 | 3.51 | 1.03 | 2.70 |
| 1500 | 1.25 | 3.06 | 1.20 | 2.91 | 1.20 | 2.87 | 1.15 | 2.78 | 1.15 | 2.56 | 1.10 | 2.53 | 1.05 | 2.40 |
| 2000 | 1.00 | 2.00 | 0.93 | 1.98 | 0.92 | 1.75 | 0.90 | 1.51 | 0.88 | 1.26 | 0.85 | 1.02 | 0.83 | 0.77 |
| 2500 | 1.05 | 2.56 | 1.00 | 2.52 | 0.90 | 2.39 | 0.89 | 2.36 | 0.80 | 2.32 | 0.77 | 2.20 | 0.73 | 2.15 |
| 3000 | 1.03 | 2.67 | 1.01 | 2.57 | 0.98 | 2.52 | 0.96 | 2.41 | 0.88 | 2.34 | 0.88 | 2.16 | 0.78 | 2.05 |

Table 4 AUC Experiment results with different hidden size for SVC2004 and SUSIG

| Iteration | 100 | | 200 | | 300 | | 400 | | 500 | | 600 | | 700 | |
|---|---|---|---|---|---|---|---|---|---|---|---|---|---|---|
| Hidden size | SVC | SUSIG | SVC | SUSIG | SVC | SUSIG | SVC | SUSIG | SVC | SUSIG | SVC | SUSIG | SVC | SUSIG |
| 500 | 0.991 | 0.980 | 0.992 | 0.980 | 0.992 | 0.980 | 0.992 | 0.981 | 0.993 | 0.982 | 0.993 | 0.982 | 0.993 | 0.988 |
| 1000 | 0.993 | 0.981 | 0.993 | 0.983 | 0.994 | 0.984 | 0.994 | 0.986 | 0.994 | 0.986 | 0.994 | 0.987 | 0.994 | 0.990 |
| 1500 | 0.993 | 0.989 | 0.993 | 0.989 | 0.994 | 0.989 | 0.994 | 0.989 | 0.994 | 0.990 | 0.995 | 0.990 | 0.995 | 0.991 |
| 2000 | 0.995 | 0.992 | 0.995 | 0.992 | 0.995 | 0.992 | 0.995 | 0.993 | 0.995 | 0.993 | 0.996 | 0.994 | 0.996 | 0.995 |
| 2500 | 0.994 | 0.989 | 0.995 | 0.990 | 0.995 | 0.991 | 0.995 | 0.991 | 0.995 | 0.991 | 0.996 | 0.991 | 0.996 | 0.992 |
| 3000 | 0.995 | 0.990 | 0.995 | 0.990 | 0.995 | 0.990 | 0.995 | 0.991 | 0.995 | 0.991 | 0.995 | 0.991 | 0.996 | 0.992 |

As a comparison between the proposed system and other approaches, verification protocols must be similar. Based on random and skilled forgery verification protocol [25, 26], 25 percent of each users' genuine signatures have been used for training to create the user model. The remaining 75 percent of users' genuine signatures, all of the skilled forgery signatures of his/her and all of the genuine signatures of other users have been used for testing based on a local threshold for each user. For evaluating the proposed method, multiple classifiers have been tested based on authors' previous work [29]. These classifiers are available in Matlab open source Data Description toolbox[4] (dd_tools). This toolbox has the ability of obtaining optimal coefficients for classifiers. Finally, based on experimental results achieved, Gaussian classifier has been used.

The results of proposed method in comparison with state-of-the-art methods for two standard benchmarks (SVC2004 and SUSIG) are shown in Table 5 and Table 6.

Table 5 Different online signature verification methods for SVC2004

| Method | EER (%) |
|---|---|
| Gruber, et al. [22] | 6.84 |
| Mohammadi and Faez [13] | 6.33 |
| Barkoula, et al. [9] | 5.33 |
| Yahyatabar, et al. [11] | 4.58 |
| Yeung, et al. [25] | 2.89 |
| Ansari, et al. [6] | 1.65 |
| Fayyaz, et al. [29] | 2.15 |
| **Proposed Method** | **0.83** |

Table 6 Different online signature verification methods for SUSIG

| Method | EER (%) |
|---|---|
| Khalil, et al. [30] | 3.06 |
| Napa and Memon [14] | 2.91 |
| Kholmatov and Yanikoglu [26] | 2.10 |
| Ibrahim, et al. [31] | 1.59 |
| Ansari, et al. [6] | 1.23 |
| **Proposed Method** | **0.77** |

These tables indicate that proposed method have the best performance in comparison with competing algorithms. This method's EER on SVC dataset is 0.83 percent, where the next best method is 1.65 percent reported for the method Ansari, et al. [6]. This verification system is 0.82 percent better than the otherwise best result. On SUSIG benchmark, implemented method's EER is equal to 0.77 percent as it is 0.46 percent better than the next best method.

Table 5 and 6 illustrate that in contrast to all reported methods, the results on two datasets are very close (0.06 percent difference in EER). This similarly is related that proposed method is dataset invariant.

[4] Available at http://www.prtools.org

## VI. Conclusions and Future Work

In this paper, a new approach has been introduced based on Self-thought learning to verify the signatures. As it can be inferred from experimental results and inherited properties of Self-thought learning, the proposed system is independent from specific benchmarks, which means that it is signature shape invariant.

The features, which are used to verify the signatures, have been extracted from ATVS dataset by using a sparse autoencoder with one hidden layer. By applying convolution and pooling methods, system has achieved pooled convolved features to verify the signatures. In addition, one-class classifier has been applied as it models the signatures of each user.

To compare with similar works, two standard benchmarks have used which are named as SVC and SUSIG datasets. Our results have shown superiority on both datasets. The features have been used in this paper can be used in other benchmarks, as this is the main component of the method proposed in this paper.

This method has proved its ability to extract the best set of features in problems that need to define hand-crafted features. Therefore, it can be used in a wide range of machine learning problems. As a future work, this method can be tested on offline signatures. In addition, the impact of deep convolutional networks can be tested on both online and offline signature datasets.

## References

[1] D. Impedovo and G. Pirlo, "Automatic Signature Verification: The State of the Art," *Systems, Man, and Cybernetics, Part C: Applications and Reviews, IEEE Transactions on,* vol. 38, pp. 609-635, 2008.

[2] D. Impedovo, G. Pirlo, and R. Plamondon, "Handwritten Signature Verification: New Advancements and Open Issues," in *Frontiers in Handwriting Recognition (ICFHR), 2012 International Conference on*, 2012, pp. 367-372.

[3] D. Menotti, G. Chiachia, A. Pinto, W. Schwartz, H. Pedrini, A. Falcao, *et al.*, "Deep Representations for Iris, Face, and Fingerprint Spoofing Detection," *Information Forensics and Security, IEEE Transactions on,* vol. 10, 2015.

[4] M. Faundez-Zanuy, "On-line signature recognition based on VQ-DTW," *Pattern Recognition,* vol. 40, pp. 981-992, 3// 2007.

[5] S. Rashidi, A. Fallah, and F. Towhidkhah, "Authentication based on signature verification using position, velocity, acceleration and Jerk signals," in *Information Security and Cryptology (ISCISC), 2012 9th International ISC Conference on*, 2012, pp. 26-31.

[6] A. Q. Ansari, M. Hanmandlu, J. Kour, and A. K. Singh, "Online signature verification using segment-level fuzzy modelling," *Biometrics, IET,* vol. 3, pp. 113-127, 2014.

[7] P. Jain and J. Gangrade, "Online Signature Verification Using Energy, Angle and Directional Gradient Feature with Neural Network," *International Journal of Computer Science and Information Technologies (IJCSIT),* vol. 5, pp. 211-216, 2014.

[8] J. Fierrez, J. Ortega-Garcia, D. Ramos, and J. Gonzalez-Rodriguez, "HMM-based on-line signature verification: Feature extraction and signature modeling," *Pattern Recognition Letters,* vol. 28, pp. 2325-2334, 12/1/ 2007.

[9] K. Barkoula, G. Economou, and S. Fotopoulos, "Online signature verification based on signatures turning angle representation using longest common subsequence matching," *International Journal on Document Analysis and Recognition (IJDAR),* vol. 16, pp. 261-272, 2013/09/01 2013.

[10] Z. Zhang, K. Wang, and Y. Wang, "A Survey of On-line Signature Verification," in *Biometric Recognition*. vol. 7098, Z. Sun, J. Lai,

X. Chen, and T. Tan, Eds., ed: Springer Berlin Heidelberg, 2011, pp. 141-149.

[11] M. E. Yahyatabar, Y. Baleghi, and M. R. Karami, "Online signature verification: A Persian-language specific approach," in *Electrical Engineering (ICEE), 2013 21st Iranian Conference on*, 2013, pp. 1-6.

[12] M. J. Alhaddad, D. Mohamad, and A. M. Ahsan, "Online Signature Verification Using Probablistic Modeling and Neural Network," in *Engineering and Technology (S-CET), 2012 Spring Congress on*, 2012, pp. 1-5.

[13] M. H. Mohammadi and K. Faez, "Matching between Important Points using Dynamic Time Warping for Online Signature Verification," *Cyber Journals: Multidisciplinary Journals in Science and Technology, Journal of Selected Areas in Bioinformatics (JBIO),* 2012.

[14] S.-B. Napa and N. Memon, "Online Signature Verification on Mobile Devices," *Information Forensics and Security, IEEE Transactions on,* vol. 9, pp. 933-947, 2014.

[15] A. Reza, H. Lim, and M. Alam, "An Efficient Online Signature Verification Scheme Using Dynamic Programming of String Matching," in *Convergence and Hybrid Information Technology*. vol. 6935, G. Lee, D. Howard, and D. Ślęzak, Eds., ed: Springer Berlin Heidelberg, 2011, pp. 590-597.

[16] R. Wang, C. Han, Y. Wu, and T. Guo, "Fingerprint Classification Based on Depth Neural Network," *The Computing Research Repository (CoRR),* vol. September 2014, 2014.

[17] M. R. P. Souza, G. D. C. Cavalcanti, and R. Tsang Ing, "Off-line Signature Verification: An Approach Based on Combining Distances and One-class Classifiers," in *Tools with Artificial Intelligence (ICTAI), 2010 22nd IEEE International Conference on*, 2010, pp. 7-11.

[18] A. Fallah, M. Jamaati, and A. Soleamani, "A new online signature verification system based on combining Mellin transform, MFCC and neural network," *Digital Signal Processing,* vol. 21, pp. 404-416, 3// 2011.

[19] V. Iranmanesh, S. M. S. Ahmad, W. A. W. Adnan, S. Yussof, O. A. Arigbabu, and F. L. Malallah, "Online Handwritten Signature Verification Using Neural Network Classifier Based on Principal Component Analysis," *The Scientific World Journal,* vol. 2014, 2014.

[20] K. Cpałka, M. Zalasiński, and L. Rutkowski, "New method for the on-line signature verification based on horizontal partitioning," *Pattern Recognition,* vol. 47, pp. 2652–2661, 2014.

[21] M. Lopez-Garcia, R. Ramos-Lara, O. Miguel-Hurtado, and E. Canto-Navarro, "Embedded System for Biometric Online Signature Verification," *Industrial Informatics, IEEE Transactions on,* vol. 10, pp. 491-501, 2014.

[22] C. Gruber, T. Gruber, S. Krinninger, and B. Sick, "Online Signature Verification With Support Vector Machines Based on LCSS Kernel Functions," *Systems, Man, and Cybernetics, Part B: Cybernetics, IEEE Transactions on,* vol. 40, pp. 1088-1100, 2010.

[23] H. Song and S.-Y. Lee, "Hierarchical Representation Using NMF," in *Neural Information Processing*. vol. 8226, M. Lee, A. Hirose, Z.-G. Hou, and R. Kil, Eds., ed: Springer Berlin Heidelberg, 2013, pp. 466-473.

[24] R. Raina, A. Battle, H. Lee, B. Packer, and A. Y. Ng, "Self-taught learning: transfer learning from unlabeled data," presented at the Proceedings of the 24th international conference on Machine learning, Corvalis, Oregon, USA, 2007.

[25] D.-Y. Yeung, H. Chang, Y. Xiong, S. George, R. Kashi, T. Matsumoto, *et al.*, "SVC2004: First International Signature Verification Competition," in *Biometric Authentication*. vol. 3072, D. Zhang and A. Jain, Eds., ed: Springer Berlin Heidelberg, 2004, pp. 16-22.

[26] A. Kholmatov and B. Yanikoglu, "SUSIG: an on-line signature database, associated protocols and benchmark results," *Pattern Analysis and Applications,* vol. 12, pp. 227-236, 2009/09/01 2009.

[27] J. Galbally, j. Plamondon, J. Fierrez, and J. Ortega-Garcia, "Synthetic on-line signature generation. Part I: Methodology and algorithms," *Pattern Recognition,* vol. 45, pp. 2610-2621, 2012.

[28] J. Galbally, J. Fierrez, J. Ortega-Garcia, and j. Plamondon, "Synthetic on-line signature generation. Part II: Experimental validation," *Pattern Recognition,* vol. 45, pp. 2622-2632, 2012.

[29] M. Fayyaz, M. H. Saffar, M. Sabokrou, M. Hoseini, and M. Fathy, "Online Signature Verification Based on Feature Representation," in *International Symposium on Artificial Intelligence and Signal Processing*, Iran, Mashhad, 2015.

[30] M. I. Khalil, M. Moustafa, and H. M. Abbas, "Enhanced DTW based on-line signature verification," in *Image Processing (ICIP), 2009 16th IEEE International Conference on*, 2009, pp. 2713-2716.

[31] M. T. Ibrahim, M. Kyan, and G. Ling, "On-line signature verification using global features," in *Electrical and Computer Engineering, 2009. CCECE '09. Canadian Conference on*, 2009, pp. 682-685.